# Automatic Myocardial Segmentation by Using A Deep Learning Network in Cardiac MRI

Ariel H. Curiale*§, Flavio D. Colavecchia‡¶, Pablo Kaluza*, Roberto A. Isoardi† and German Mato‡§
*CONICET - FCEN, Universidad Nacional de Cuyo, Padre Jorge Contreras 1300, 5500 Mendoza, Argentina
†CNEA - Fundación Escuela Medicina Nuclear, Garibaldi 405, 5500, Mendoza, Argentina
‡Comisión Nacional de Energía Atómica - CONICET
§Departamento de Física Médica, Centro Atómico Bariloche,
¶Centro Integral de Medicina Nuclear y Radioterapia Bariloche,
Avenida Bustillo 9500, 8400 S. C. de Bariloche, Río Negro, Argentina
Email: see http://www.curiale.com.ar

*Abstract*—Cardiac function is of paramount importance for both prognosis and treatment of different pathologies such as mitral regurgitation, ischemia, dyssynchrony and myocarditis. Cardiac behavior is determined by structural and functional features. In both cases, the analysis of medical imaging studies requires to detect and segment the myocardium. Nowadays, magnetic resonance imaging (MRI) is one of the most relevant and accurate non-invasive diagnostic tools for cardiac structure and function.

In this work we propose to use a deep learning technique to assist the automatization of myocardial segmentation in cardiac MRI. We present several improvements to previous works in this paper: we propose to use the Jaccard distance as optimization objective function, we integrate a residual learning strategy into the code, and we introduce a batch normalization layer to train the fully convolutional neural network. Our results demonstrate that this architecture outperforms previous approaches based on a similar network architecture, and that provides a suitable approach for myocardial segmentation. Our benchmark shows that the automatic myocardial segmentation takes less than 22 seg. for a volume of 128 x 128 x 13 pixels in a 3.1 GHz intel core i7.

## I. INTRODUCTION

Quantitative characterization of cardiac function is highly desirable for both prognosis and treatment of different pathologies such as mitral regurgitation, ischemia, dyssynchrony and myocarditis [1], [2], [3], [4], [5]. In particular, cardiac function is characterized by structural and functional features. These features can be classified as global or regional, according to the type of information they provide. One of the most relevant global structural features are the left ventricular mass (LVM), the left ventricular volume (LV) and the ejection fraction (EF), which is directly derived from the LV at end-diastole (ED) and end-systole (ES). Besides, the most relevant functional features are those derived from myocardial deformation such as strain and strain rate. Functional information is commonly presented according to the 17-segment model proposed by the American Heart Association (AHA) [6]. In both cases, it is necessary to detect and segment the myocardium of the left ventricle.

Magnetic resonance imaging (MRI) is one of the most important and accurate non-invasive diagnostic tools for imaging of cardiac structure and function [7]. It is usually considered as the gold standard for ventricular volume quantification [8]. Chamber segmentation from cardiac MRI datasets is an essential step for quantification of global features, which involves the estimation of ventricular volume, ejection fraction, mass and wall thickness, among others. Manual delineation by experts is currently the standard clinical practice for performing chamber segmentation. However, and despite the efforts of researchers and medical vendors, global quantification and volumetric analysis still remain time consuming tasks which heavily rely on user interaction. For example, the time required to accurately extract the most common volumetric indices for the left ventricle for a single patient is around 10 minutes [9]. Thus, there is still a significant need for tools allowing automatic 3D quantification. The main goal of this work is to provide a suitable and accurate automatic myocardial segmentation approach for cardiac MRI based on deep convolutional neural networks [10], [11].

Many different techniques have been applied in the past to treat the problem of automatic segmentation of the left ventricle. Most of these approaches depend on proper initialization step [12]. Many model-based left ventricle segmentation approaches have been studied[13]: active contours [14], level-set method [15], active shape models and active appearance models [16], [17]. Unfortunately, all of them have a strong dependence on the initial model placement. In general, improper initialization leads to unwanted local minima of the objective function.

In contrast, non model-based approaches to the problem are expected to have a more robust behavior. Among these approaches are Convolutional Neural Networks (CNNs), which combine insights from the structure of receptive fields of the visual system with modern techniques of machine learning [18]. Several recent studies in computer vision and pattern recognition highlight the capabilities of these networks to solve challenging tasks such as classification, segmentation and object detection, achieving state-of-the-art performances [19]. Fully convolutional networks trained end-to-end have been recently used for medical image segmentation [20], [21]. These models, which serve as an inspiration for our work, employ different types of network architectures and were trained to generate a segmentation mask that delineates the structures of interest in the image.

One of the main disadvantages of *deep neural networks*, such as CNN, is that its training is complicated by the fact that the distribution of each layer's inputs changes along this process, as the parameters of the previous layers change. This phenomenon is called internal covariate shift [22]. As the networks start to converge, a degradation problem occurs: with increasing network depth, accuracy gets saturated. Unexpectedly, such degradation is not caused by overfitting, and adding more layers to a suitably deep model leads to higher training error, as reported [23], [24]. To overcome this problem the technique of batch normalization has been proposed [25]. This procedure involves the evaluation of the statistical properties of the neural activations that are present for a given batch of data in order to normalize the inputs to any layer to obtain some desired objective (such as zero mean and unit variance). At the same time, the architecture is modified in order to prevent loss of information that could arise from the normalization. This technique allows to use much higher learning rates and also acts as regularizer, in some cases eliminating the need for dropout [26].

Another recent performance improvement of deep neural networks has been achieved by reformulating the layers as learning residual functions with reference to the layer inputs, instead of learning unreferenced functions [27]. In other words, the parameters to be determined at a given stage generate only the difference (or residual) between the objective function to be learned and some fixed function such as the identity. It was empirically found that this approach gives rise to networks that are easier to optimize, which can also gain accuracy from considerably increased depth [27].

In this work, we analyze the use of these techniques in the U-net architecture to segment the myocardium of the left ventricle, similarly as proposed in [21] for studying the prostate. Unlike previous works, we propose to use the Jaccard distance [28] as optimization objective and a batch normalization layer for the training of the CNN. Results demonstrate that the proposed model outperforms previous approaches based on the U-net architecture and provides a suitable approach for myocardial segmentation.

The paper is structured as follows: in Section 2 the U-net network with residual learning and batch normalization is introduced. Additionally, the training strategy and the objective function used for training the CNN is presented. In Section 3, the evaluation of the proposed method in cardiac MRI is presented. Finally, we present the conclusions in Section 4.

## II. MATERIALS AND METHODS

*Materials*

The present network for myocardial segmentation was trained and evaluated with the Sunnybrook Cardiac Dataset (SCD), also known as the 2009 Cardiac MR Left Ventricle Segmentation Challenge data [29]. The dataset consists of 45 cine-MRI images from a mix of patients and pathologies: healthy, hypertrophy, heart failure with infarction and heart failure without infarction. A subset of this dataset was first used in the automated myocardium segmentation challenge from short-axis MRI, held by a MICCAI workshop in 2009. The whole complete dataset is now available in the Cardiac Atlas Project database with public domain license[1].

The Sunnybrook Cardiac Dataset [29] contains 45 cardiac cine-MR datasets with expert contours, as part of the clinical routine. Cine steady state free precession (SSFP) MR short axis (SAX) images were obtained with a 1.5T GE Signa MRI. All the images were obtained during 10-15 second breath-holds with a temporal resolution of 20 cardiac phases over the heart cycle, and scanned from the ED phase. In these SAX MR acquisitions, endocardial and epicardial contours were drawn by an experienced cardiologist in all slices at ED and ES phases. All the contours were confirmed by another cardiologist. The manual segmentations will be used as the ground truth for evaluation purposes. Additionally, a ROI of 128 x 128 pixels with a spatial resolution of 1.36 x 1.36 mm in the short-axis around the heart was manually selected. The ROI size was defined by an expert to ensure that the entire left ventricle was included.

*Network architecture*

The modified U-net system that we use for myocardial segmentation, follows a typical encode-decode network architecture. In this architecture, depicted in Fig. 1, the network learns how to encode information about features presented in the training set (left side on Fig. 1). Then, in the decode path the network learns about the image reconstruction process from the encoded features learned previously. The specific feature of the U-net architecture lies on the concatenation between the output of the encode path, for each level, and the input of the decoding path (denoted as big gray arrows on Fig. 1). These concatenations provide the ability to localize high spatial resolution features to the *fully convolutional network*, thus, generating a more precise output based on this information. As mentioned in [20] this strategy allows the seamless segmentation of large images by an overlap-tile strategy. In order to predict the pixels in the border region of the image, the missing context is extrapolated by the concatenation from the encoding path.

The encoding path consists of the repeated application of two 3 x 3 convolutions, each followed by a rectified linear unit (ReLU) as it was originally proposed in [20]. After performing the convolution, a batch normalization is carried out to improve the accuracy and reduce learning time. Then, a residual learning is introduced just before performing the 2 x 2 max pooling operation with stride 2 for downsampling, in a similar way as it was proposed in [21]. At each downsampling step we double the number of feature channels, that is initially set to 64.

Every step in the decoding path can be seen as the mirrored step of the encode path, i.e. each step in the decoding path consists of an upsampling of the feature map followed by a 2 x 2 convolution ("up-convolution") that halves the number of feature channels, a concatenation with the corresponding feature map from the encoding path, and two 3 x 3 convolutions, each followed by a ReLU and a batch normalization. Finally, a residual learning is introduced to get the input of the next level. At the final layer, a 1 x 1 convolution is carried out to map the 64 feature maps to the two classes used for the myocardial segmentation (myocardium and background). The output of the last layer, after soft-max non-linear function, represents the likelihood of a pixel belongs to the myocardium of the left ventricle. Indeed, only those voxels with higher likelihood ($> 0.5$) are considered as part of the left ventricle tissue.

---

[1]http://www.cardiacatlas.org/studies/sunnybrook-cardiac-data

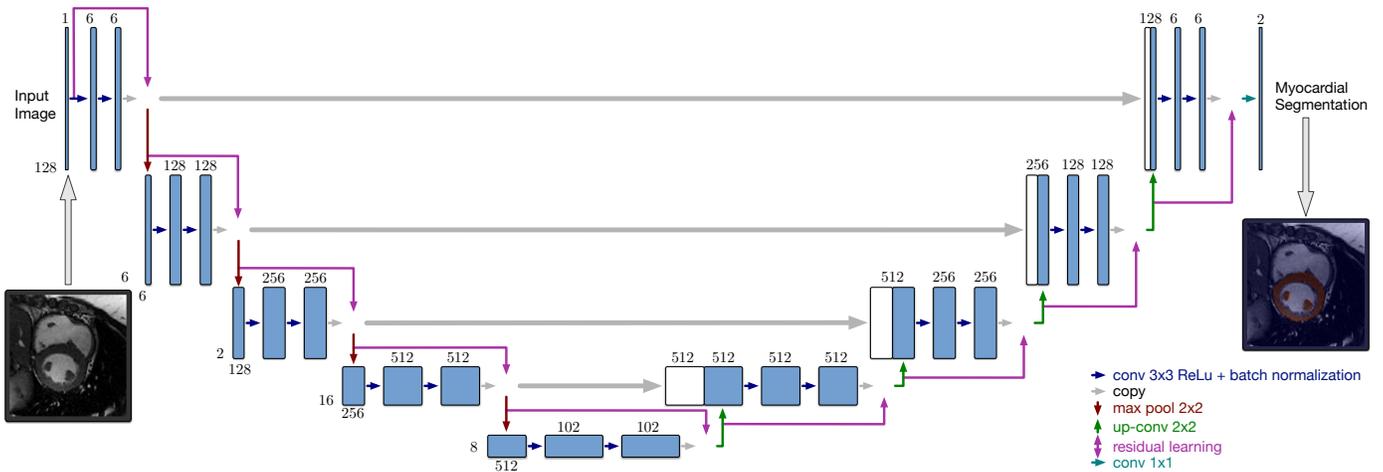

Figure 1: Network Architecture proposed for myocardial segmentation in cardiac MRI. The number of channels is denoted on top of the box and the input layer dimension is provided at the lower left edge of the box according to the U-net diagram [20]. The arrows denote the different operations according to the legend.

*Training*

Half of the images and their corresponding myocardial segmentations were manually cropped into 128 x 28 pixels with a spatial resolution of 1.36 x 1.36 mm in the short-axis around the heart. This ROI size was defined by an expert to ensure that the entire left ventricle was included for all the images. Then, the cropped images and their corresponding myocardial segmentations were used to train the network with the stochastic gradient descent (Adaptive Moment Estimation) implementation of Keras [30]. In the proposed U-net we use the Jaccard distance as objective function instead of the Dice's coefficient commonly used in image processing. The Jaccard distance is defined as follows:

$$\mathrm{Jd}(\mathrm{myo}^{\mathrm{pred}}, \mathrm{myo}^{\mathrm{true}}) = 1 - \frac{\sum_i \mathrm{myo}_i^{\mathrm{pred}} * \mathrm{myo}_i^{\mathrm{true}}}{\sum_i \mathrm{myo}_i^{\mathrm{pred}} + \sum_i \mathrm{myo}_i^{\mathrm{true}} - \sum_i \mathrm{myo}_i^{\mathrm{pred}} * \mathrm{myo}_i^{\mathrm{true}}},$$

where $\mathrm{myo}^{\mathrm{pred}}$ and $\mathrm{myo}^{\mathrm{true}}$ are the myocardial segmentation prediction and the ground truth segmentation, and the $\sum_i \mathrm{myo}_i$ refers to the pixel sum over the entire segmentation (prediction or ground truth).

Annotated medical information like myocardial classification is not easy to obtain due to the fact that one or more experts are required to manually trace a reliable ground truth of the myocardial classification. So, in this work it was necessary to augment the original training dataset in order to increase the examples. Also, data augmentation is essential to teach the network the desired invariance and robustness properties. Heterogeneity in the cardiac MRI dataset is needed to teach the network some shift and rotation invariance, as well as robustness to deformations. With this intention, during the every training iteration, the input of the network was randomly deformed by means of a spatial shift in a range of 10% of the image size, a rotation in a range of $10°$ in the short axis, a zoom in a range of 2x or by using a gaussian deformation field ($\mu = 10$ and $\sigma = 20$) and a B-spline interpolation. The data augmentation has been performed on-the-fly to reduce the data storage and achieve a total of 5500 new images by epoch.

## III. RESULTS

Three sets of experiments are conducted to evaluate the proposed methodology on the Sunnybrooks dataset [29]. In the first two experiments, the papillary muscles (PM) were excluded on both, the automatic and manual segmentation. So, the proposed method will avoid to detect the PM as part of the myocardial tissue.

In the first experiment, the proposed objective function loss, batch normalization and residual learning were evaluated on the classic U-net architecture with respect to the commonly used Dice's coefficient. Empirical results show that the Jaccard distances ($0.6 \pm 0.1$ mean accuracy Dice's value) outperform the Dice's coefficient ($0.58 \pm 0.1$ mean accuracy Dice's value) when it is used as the objective loss function (Fig. 2a). Moreover, Fig. 2a shows that this improvement (Jaccard distance vs. Dice's coefficient) is uncorrelated with the use of batch normalization and residual learning.

As results show on Fig. 2a and Table I, the use of the batch normalization and residual learning strategies are of paramount importance to achieve an accurate myocardial segmentation.

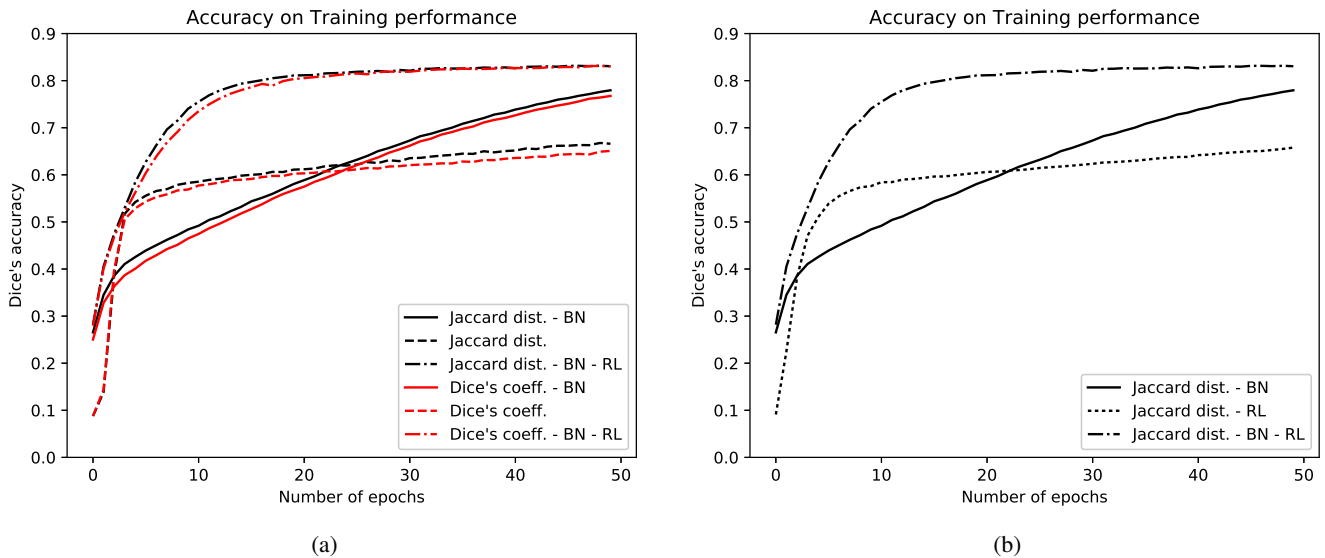

Figure 2: Accuracy on training performance for 50 training iteration over the entire test dataset (50 epochs) with a shrink factor of 2 (i.e. the original input size was reduced by a factor of 2). In the picture is depicted the accuracy performance between Dice's coefficient and Jaccard distance (JD) objective loss function, batch normalization (BN) and residual learning (RL) on the U-net architecture. For clarity in (b) it is showed the effect of BN and RL only for the Jaccard distance loss function.

Furthermore, the combination of both strategies shows the higher myocardial segmentation accuracy. However, if the training accuracy with the strategies of batch normalization and residual learning is analyzed, we can conclude that the contribution of the batch normalization is higher than contribution of the residual learning (Fig. 2b). Also, we can conclude that the training accuracy of the residual learning is comparable to the basic U-net for our architecture (i.e. 5 levels, 2 conv. operation by level and the same number of features by level). Nonetheless, we strongly recommend to use both strategies together due to the accuracy achieved by them working together is remarkable higher than the individual contribution of each one as it can be seen in Fig. 2b.

| SF | Architecture | Dice's acc. | MSE | MAE |
|---|---|---|---|---|
| 2 | U-net - Dice's | 0.7306 | 0.0254 | 0.0321 |
|   | U-net - JD | 0.7308 | 0.0256 | 0.0321 |
|   | U-net - BN - JD | 0.8418 | 0.0126 | 0.0229 |
|   | U-net - BN - RL - JD | **0.8799** | 0.0114 | 0.0181 |
| 1 | U-net - BN - RL - JD | **0.9001** | 0.0093 | 0.0094 |

Table I: Myocardial segmentation accuracy for each convolutional neural network studied on the Sunnybrooks dataset for 50 training iteration over the entire test dataset (50 epochs). SF: shrink factor with respect to the original image (i.e. SF=2 means that the original input size was reduced by a factor of 2). MSE: mean squeared error. MAE: mean absolute error. JD: Jaccard distance. BN: Batch normalization. RL: Residual learning.

The second experiment shows (Fig. 3) that our CNN for myocardial segmentation in cardiac MRI reaches suitable accuracy in a few training iterations. i.e. in 30 epochs. In this case, the present CCN reaches a value of 0.9 and 0.0093 for the Dice's coefficient and a mean squared error, respectively, which is comparable to manual segmentation.

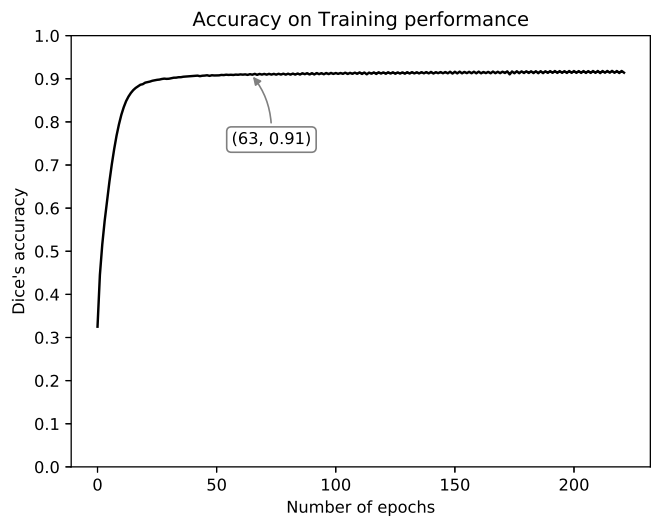

Figure 3: Accuracy on training performance of the proposed U-net architecture for myocardial segmentation in cardiac MRI.

Finally, as it was expected, the third experiment revealed that the accuracy tend to decrease when the PM were taking into account as part of the myocardial tissue, since it makes the myocardial tissue segmentation harder. However, the proposed approach seems to be not affected and reaches a similar accuracy to the previous experiment. We obtained 0.89 for the Dice's coefficient and a mean squared error of 0.01.

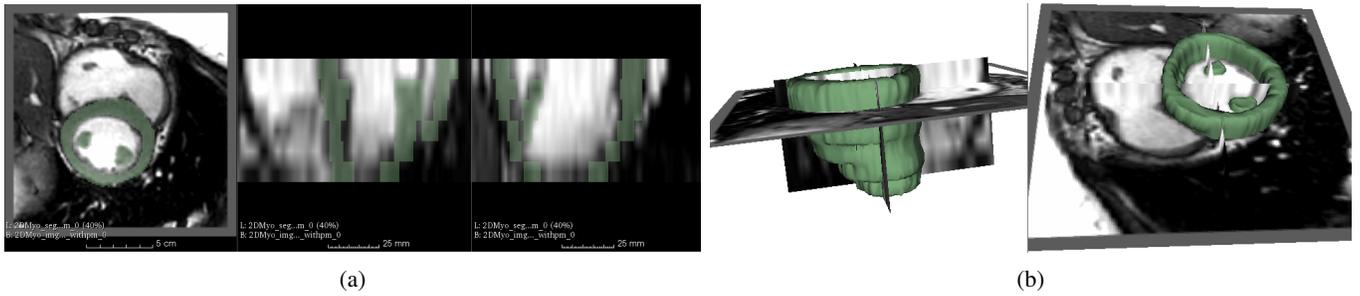

Figure 4: Example of myocardial segmentation for the proposed convolutional neural network. (a) The proposed myocardial segmentation is presented in three orthogonal views. (b) 3D view of the proposed myocardial segmentation.

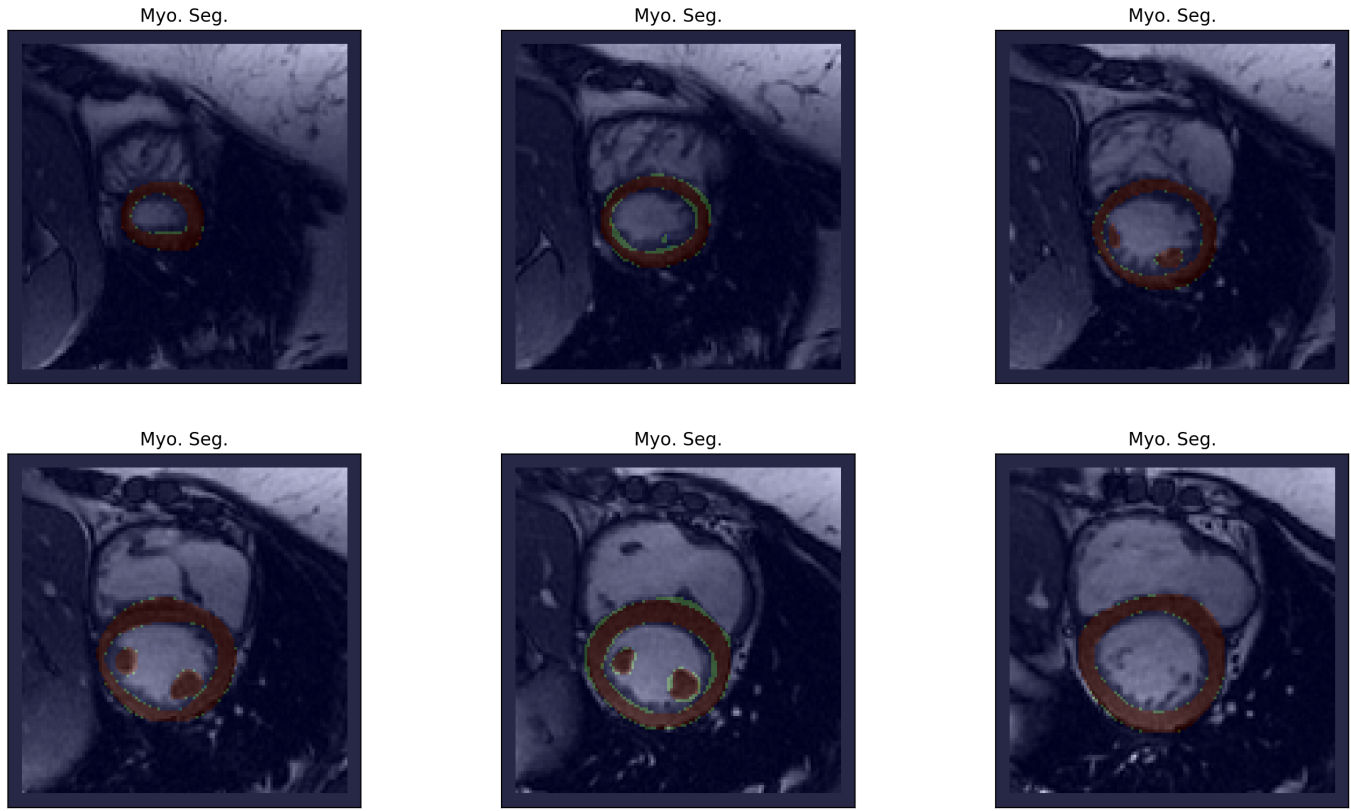

Figure 5: Qualitative results for a patient on Sunnybrook dataset. Different short axis slices are plotted from apex (upper-left) to base (down-right) of the left ventricle. In brown it is depicts those voxels where the manual myocardial segmentation and the proposed automatic segmentation overlaps, otherwise, the voxels are depicted in light-green.

An example of the automatic 3D myocardial segmentation can be seen in Fig. 4. In particular, Fig. 4a shows the myocardial segmentation in three orthogonal views and Fig. 4b shows the segmentation in two 3D views. Qualitative details of the this Automatic 3D myocardial segmentation (Fig. 4) can be seen in several slices in Fig. 5. They show that the proposed myocardial segmentation provides a suitable approach for myocardial segmentation in cardiac MRI. In yellow it is depicts those voxels where the manual myocardial segmentation and the proposed automatic segmentation overlaps, otherwise, the voxels are depicted in magenta. As it can be seen only a reduced number of voxels correspond to a non-overlapping classification.

IV. CONCLUSION

In this paper, we have proposed an automatic 3D myocardial segmentation approach for cardiac MRI by using a deep learning network. Unlike previous approaches, our method makes use of the Jaccard distance as objective loss function, a batch normalization and residual learning strategies to provide a suitable approach for myocardial segmentation. Quantitative and qualitative results show that the proposed approach presents a high potential for being used to estimate different structural and functional features for both prognosis and treatment of different pathologies. Thanks to real time data augmentation with free form deformations, it only need very few annotated images (22 cardiac MRI) and has a very reasonable training time of only 8.5 hours on a NVidia Tesla C2070 (6 GB) for reaching a suitable accuracy of 0.9 Dice's coefficient and 0.0093 mean squared error in 30 epochs.

The segmentation and tracking of different cardiac structures, such as the left ventricle, have an important role in the treatment of different pathologies. However, accurate and reliable measurements for volumetric analysis and functional assessment heavily rely on user interaction. The results obtained in this work can be efficiently used for estimating different structural (left ventricle mass, left ventricle volume and ejection fraction among others) features. Additionally, this work leads to extensions for automatic detection and tracking of the right and left ventricle. Myocardial motion is useful in the evaluation of regional cardiac functions such as the strain and strain rate.

ACKNOWLEDGMENTS

This work was partially supported by Consejo Nacional de Investigaciones Científicas y Técnicas (CONICET) and by grants M028-2016 SECTyP and L017-2016 SECTyP, Universidad Nacional de Cuyo, Argentina; PICT 2016-0091, Agencia Nacional de Promoción Científica y Tecnológica, Argentina. German Mato and Pablo Kaluza acknowledge CONICET for the grant PIP 112 201301 00256 and PIP 112 201501 00013 respectively.

REFERENCES

[1] B. D. Lowes, E. A. Gill, W. T. Abraham, J.-R. Larrain, A. D. Robertson, M. R. Bristow, and E. M. Gilbert, "Effects of carvedilol on left ventricular mass, chamber geometry, and mitral regurgitation in chronic heart failure," *The American journal of cardiology*, vol. 83, no. 8, pp. 1201–1205, 1999.

[2] T. M. Koelling, K. D. Aaronson, R. J. Cody, D. S. Bach, and W. F. Armstrong, "Prognostic significance of mitral regurgitation and tricuspid regurgitation in patients with left ventricular systolic dysfunction," *American heart journal*, vol. 144, no. 3, pp. 524–529, 2002.

[3] M. G. Friedrich, U. Sechtem, J. Schulz-Menger, G. Holmvang, P. Alakija, L. T. Cooper, J. A. White, H. Abdel-Aty, M. Gutberlet, S. Prasad *et al.*, "Cardiovascular magnetic resonance in myocarditis: A jacc white paper," *Journal of the American College of Cardiology*, vol. 53, no. 17, pp. 1475–1487, 2009.

[4] T. Edvardsen, S. Urheim, H. Skulstad, K. Steine, H. Ihlen, and O. A. Smiseth, "Quantification of left ventricular systolic function by tissue doppler echocardiography," *Circulation*, vol. 105, no. 17, pp. 2071–2077, 2002.

[5] M. S. Suffoletto, K. Dohi, M. Cannesson, S. Saba, and J. Gorcsan, "Novel speckle-tracking radial strain from routine black-and-white echocardiographic images to quantify dyssynchrony and predict response to cardiac resynchronization therapy," *Circulation*, vol. 113, no. 7, pp. 960–968, 2006.

[6] M. D. Cerqueira, N. J. Weissman, V. Dilsizian, A. K. Jacobs, S. Kaul, W. K. Laskey, D. J. Pennell, J. A. Rumberger, T. Ryan, and M. S. Verani, "Standardized myocardial segmentation and nomenclature for tomographic imaging of the heart," *Circulation*, vol. 105, no. 4, pp. 539–542, Jan. 2002.

[7] J. W. Weinsaft, I. Klem, and R. M. Judd, "Mri for the assessment of myocardial viability," *Cardiology Clinics*, vol. 25, no. 1, pp. 35–56, 2007.

[8] A. L. Gerche, G. Claessen, A. Van de Bruaene, N. Pattyn, J. Van Cleemput, M. Gewillig, J. Bogaert, S. Dymarkowski, P. Claus, and H. Heidbuchel, "Cardiac mriclinical perspective," *Circulation: Cardiovascular Imaging*, vol. 6, no. 2, pp. 329–338, 2013.

[9] E. Heijman, J.-P. Aben, C. Penners, P. Niessen, R. Guillaume, G. van Eys, K. Nicolay, and G. J. Strijkers, "Evaluation of manual and automatic segmentation of the mouse heart from cine mr images," *Journal of Magnetic Resonance Imaging*, vol. 27, no. 1, pp. 86–93, 2008.

[10] A. Krizhevsky, I. Sutskever, and G. E. Hinton, "Imagenet classification with deep convolutional neural networks," in *Advances in neural information processing systems*, 2012, pp. 1097–1105.

[11] K. Simonyan and A. Zisserman, "Very deep convolutional networks for large-scale image recognition," *arXiv preprint arXiv:1409.1556*, 2014.

[12] A. K. Jain, Y. Zhong, and M.-P. Dubuisson-Jolly, "Deformable template models: A review," *Signal Proccessing*, vol. 71, no. 2, pp. 109–129, Dec. 1998.

[13] C. Petitjean and J.-N. Dacher, "A review of segmentation methods in short axis cardiac MR images," *Medical Image Analysis*, vol. 15, no. 2, pp. 169–184, Apr. 2011.

[14] M. Kass, A. Witkin, and D. Terzopoulos, "Snakes: Active contour models," *Int. J. Comput. Vision*, vol. 1, no. 4, pp. 321–331, Jan. 1998.

[15] S. Osher and J. A. Sethian, "Fronts propagating with curvature-dependent speed: Algorithms based on hamilton-jacobi formulations," *J. Comput. Phys.*, vol. 79, no. 1, pp. 12–49, Nov. 1988.

[16] T. Cootes and C. Taylor, "Active shape models: Smart snakes," in *Proc. British Machine Vision Conference*, vol. 266275. Citeseer, 1992.

[17] T. Cootes, G. Edwards, and C. Taylor, "Active appearance models," *Computer Vision and Image Understanding*, vol. 61, no. 1, pp. 39–59, 1995.

[18] Y. Lecun, L. Bottou, Y. Bengio, and P. Haffner, "Gradient-based learning applied to document recognition," *Proceedings of the IEEE*, vol. 86, no. 11, pp. 2278–2324, Nov 1998.

[19] G. Litjens, T. Kooi, B. E. Bejnordi, A. A. A. Setio, F. Ciompi, M. Ghafoorian, J. A. van der Laak, B. van Ginneken, and C. I. Sánchez, "A survey on deep learning in medical image analysis," *arXiv preprint arXiv:1702.05747*, 02 2017.

[20] O. Ronneberger, P. Fischer, and T. Brox, "U-net: Convolutional networks for biomedical image segmentation," in *Medical Image Computing and Computer-Assisted Intervention – MICCAI 2015: 18th International Conference, Munich, Germany, October 5-9, 2015, Proceedings, Part III*, N. Navab, J. Hornegger, W. M. Wells, and A. F. Frangi, Eds. Cham: Springer International Publishing, 2015, pp. 234–241.

[21] F. Milletari, N. Navab, and S. A. Ahmadi, "V-net: Fully convolutional neural networks for volumetric medical image segmentation," in *2016 Fourth International Conference on 3D Vision (3DV)*, Oct 2016, pp. 565–571.

[22] H. Shimodaira, "Improving predictive inference under covariate shift by weighting the log-likelihood function," *Journal of statistical planning and inference*, vol. 90, no. 2, pp. 227–244, 2000.

[23] K. He and J. Sun, "Convolutional neural networks at constrained time cost," in *The IEEE Conference on Computer Vision and Pattern Recognition (CVPR)*, June 2015.

[24] R. K. Srivastava, K. Greff, and J. Schmidhuber, "Highway networks," *arXiv preprint arXiv:1505.00387*, 05 2015.

[25] S. Ioffe and C. Szegedy, "Batch Normalization: Accelerating Deep Network Training by Reducing Internal Covariate Shift," *ArXiv e-prints*, Feb. 2015.

[26] N. Srivastava, G. E. Hinton, A. Krizhevsky, I. Sutskever, and R. Salakhutdinov, "Dropout: a simple way to prevent neural networks from overfitting." *Journal of Machine Learning Research*, vol. 15, no. 1, pp. 1929–1958, 2014.

[27] K. He, X. Zhang, S. Ren, and J. Sun, "Deep residual learning for image recognition," in *The IEEE Conference on Computer Vision and Pattern Recognition (CVPR)*, June 2016.

[28] R. Toldo and A. Fusiello, "Robust multiple structures estimation with j-linkage," *Computer Vision–ECCV 2008*, pp. 537–547, 2008.

[29] P. Radau, Y. Lu, K. Connelly, G. Paul, A. J. Dick, and G. A. W. GA., "Evaluation framework for algorithms segmenting short axis cardiac mri," *The MIDAS Journal -Cardiac MR Left Ventricle Segmentation Challenge*, 2009.

[30] F. Chollet, "Keras," https://github.com/fchollet/keras, 2015.